\documentclass[journal]{IEEEtran}
\usepackage{times}  
\usepackage{helvet} 
\usepackage{courier}  
\usepackage[hyphens]{url}  
\usepackage{graphicx} 
\usepackage{epsfig}
\usepackage{amsmath}
\usepackage{amssymb}
\usepackage{multirow}
\usepackage{makecell}
\usepackage{bm}
\usepackage{bigstrut}
\usepackage{float}
\renewcommand{\figurename}{Fig.}

\title{ 
Multi-Domain Joint Training for Person Re-Identification
}
\author{Lu~Yang,
        Lingqiao~Liu, 
        Yunlong~Wang,
        Peng~Wang, 
        and~Yanning~Zhang 
}

\begin{document}

\markboth{xxxx} 
{Shell \MakeLowercase{\textit{et al.}}: Multi-Domain Joint Training for Person Re-Identification }

\maketitle

\begin{abstract}
Deep learning-based person Re-IDentification (ReID) often requires a large amount of training data to achieve good performance. Thus it appears that collecting more training data from diverse environments tends to improve the ReID performance.  This paper re-examines this common belief and makes a somehow surprising observation: using more samples, i.e., training with samples from multiple datasets, does not necessarily lead to better performance by using the popular ReID models. In some cases, training with more samples may even hurt the performance of the evaluation is carried out in one of those datasets. We postulate that this phenomenon is due to the incapability of the standard network in adapting to diverse environments. To overcome this issue, we propose an approach called Domain-Camera-Sample Dynamic network (DCSD) whose parameters can be adaptive to various factors. Specifically, we consider the internal domain-related factor that can be identified from the input features, and external domain-related factors, such as domain information or camera information. Our discovery is that  training with such an adaptive model can better benefit from more training samples.
Experimental results show that our DCSD can greatly boost the performance (up to $12.3\%$) while joint training in multiple datasets.
\end{abstract}


\begin{table*}[t]
\caption{The statistics of person re-identification datasets in our experiments. ``*`` denotes that we modified the dataset by using the ground-truth bounding box annotation for our experiments rather than using the original images which were originally used for person search evaluation.}
\label{tab:datasets}
\centering
\begin{tabular}{c|ccc|ccc|ccc}
\hline \hline
\multirow{2}{*}{}  & \multicolumn{3}{c|}{Training Set} & \multicolumn{3}{c|}{Test Set (Query)} & \multicolumn{3}{c}{Test Set (Gallery)} \\ \cline{2-10} 
& ID\# & Image\# & Camera\# & ID\# & Image\# & Camera\# & ID\#  & Image\# & Camera\#\\ \hline
Market1501 & $751$ & $12,936$ & $6$ & $750$ & $3,368$ & $6$ & $751$ & $15,913$ & $6$\\
CUHK-SYSU* &  $942$ & $4,374$ & $1$ & $2,900$ & $2,900$ & $1$ & $2,900$ & $5,447$ & $1$\\
Duke & $702$ & $16,522$ & $8$ & $702$ & $2,228$ & $8$ & $1,110$ & $17,661$ & $8$\\
CUHK03 & $767$ & $7,368$ & $2$ & $700$ & $1,400$ & $2$ & $700$ & $5,328$ & $2$\\
MSMT17 & $1,041$ & $30,248$ & $15$ & $3,060$ & $11,657$ & $15$ & $3,060$ & $82,161$ & $15$\\
\hline \hline
\end{tabular}
\end{table*}

\section{Introduction}
Due to the urgent demand for public security and the increasing number of surveillance cameras on campus, parks, streets, and shopping malls, person Re-IDentification (ReID) embraces many applications in intelligent video surveillance and presents a significant challenge in computer vision. ReID, as an important application of deep metric learning, aims to retrieve the images with the same identity of queries. The common solution is to construct an embedding space such that samples with identical identities (IDs) are gathered while samples with different IDs are well separated. 
Presently, deep learning methods dominate this community, with convincing superiority against hand-crafted competitors. The development of deep convolution network \cite{krizhevsky2012imagenet,he2016deep} introduces a more powerful representation method for pedestrian images, which boosts the performance of ReID to a high level.

\begin{figure}[t]
\begin{center}
\resizebox{0.5\textwidth}{!}{
\includegraphics[width=\linewidth]{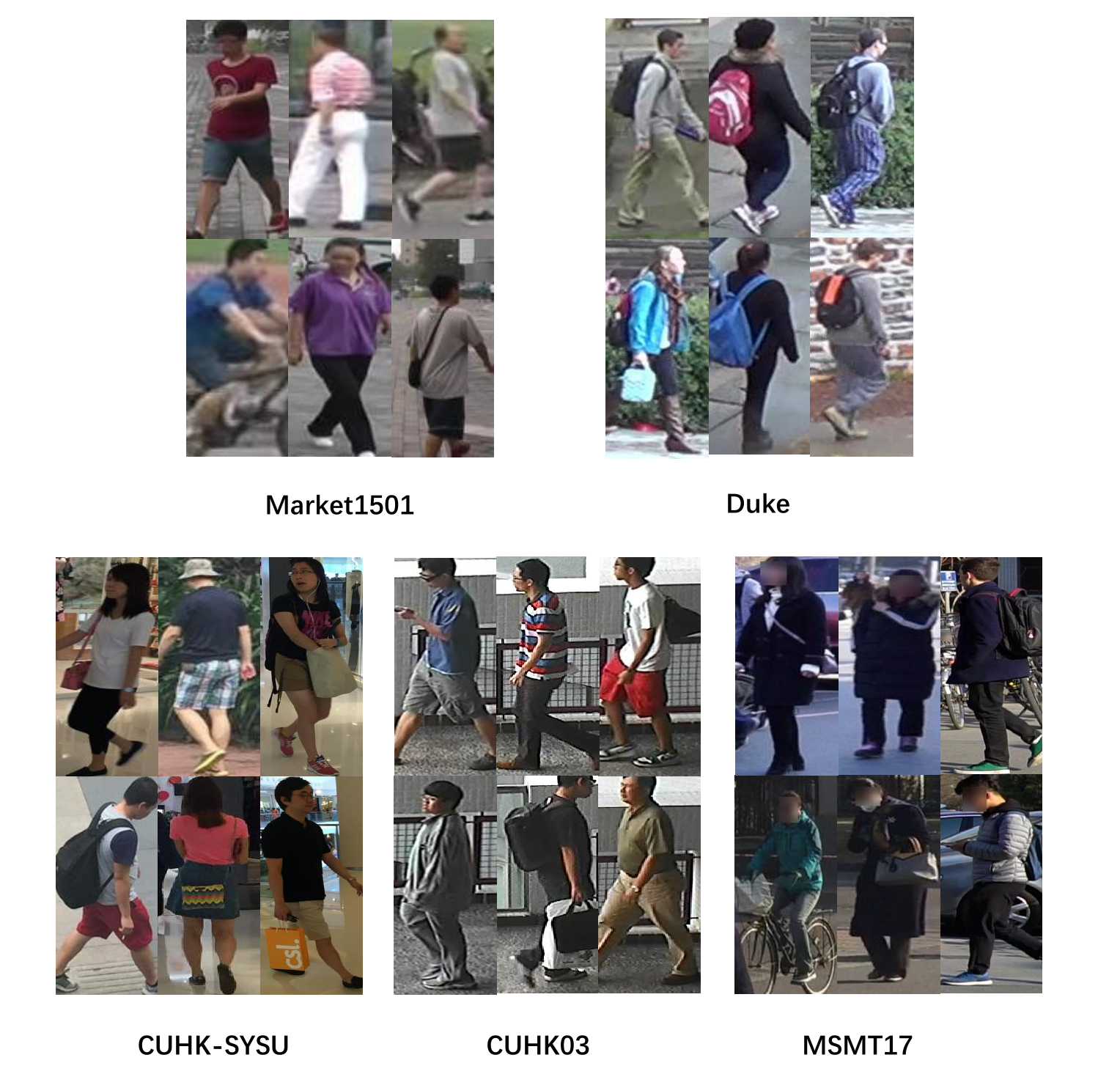}
}
\caption{Illustration of the domain gap between the person Re-Identification datasets. We can find that the people in Market1501 wear cool clothes, people in Duke often carry schoolbags, people in CUHK-SYSU wear cool clothes and often go to shopping malls, people in CUHK03 seem to often go in and out of buildings, and people in MSMT17 wear warm clothes.
}
\label{fig:domaingap}
\end{center}
\end{figure}

In real scenarios, there may be few person images in some special domains. In order to improve the performance of the models deployed in the domain, researchers may use more person images collected from other domains for joint training. Please note that the model is jointly trained on multiple domains during training, but only deployed on one of the source domains during the test phase.
Generally speaking, the performance of deep convolution network on vision tasks increases based on the volume of training data size~\cite{sun2017revisiting}. That means the more data, the better performance should be. However, as Table~\ref{tab:SOTA_person} shows, when multiple domains are jointly trained, the performance of person ReID is even worse than that of each domain independently trained. We call it \emph{domain conflict problem}.
The domain conflicts may come from three different aspects: datasets (domain), cameras, and samples.
Different person ReID datasets are collected from different regions and times, resulting in great differences in appearance. As shown in Figure~\ref{fig:domaingap}, people in CUHK-SYSU wear cool clothes and often go to shopping malls, but people in MSMT17 wear warm clothes.
\cite{Qi_2019_ICCV} highlight the presence of camera-level sub-domains as a unique characteristic in person ReID.
And some works~\cite{li2021dynamic} treat each sample as an independent micro domain.
Extracting robust feature representation is a key challenge in ReID. However, it is difficult for a traditional static network to extract robust feature representations in all domains.
Because once the traditional convolution network is trained, the network parameters are fixed.
Due to the domain conflict problem, it will be difficult for a traditional approach to deal with the data of all domains at the same time, especially when the data distribution of these domains is very different.

\begin{figure}[t]
\begin{center}
\resizebox{0.49\textwidth}{!}{
\includegraphics[width=\linewidth]{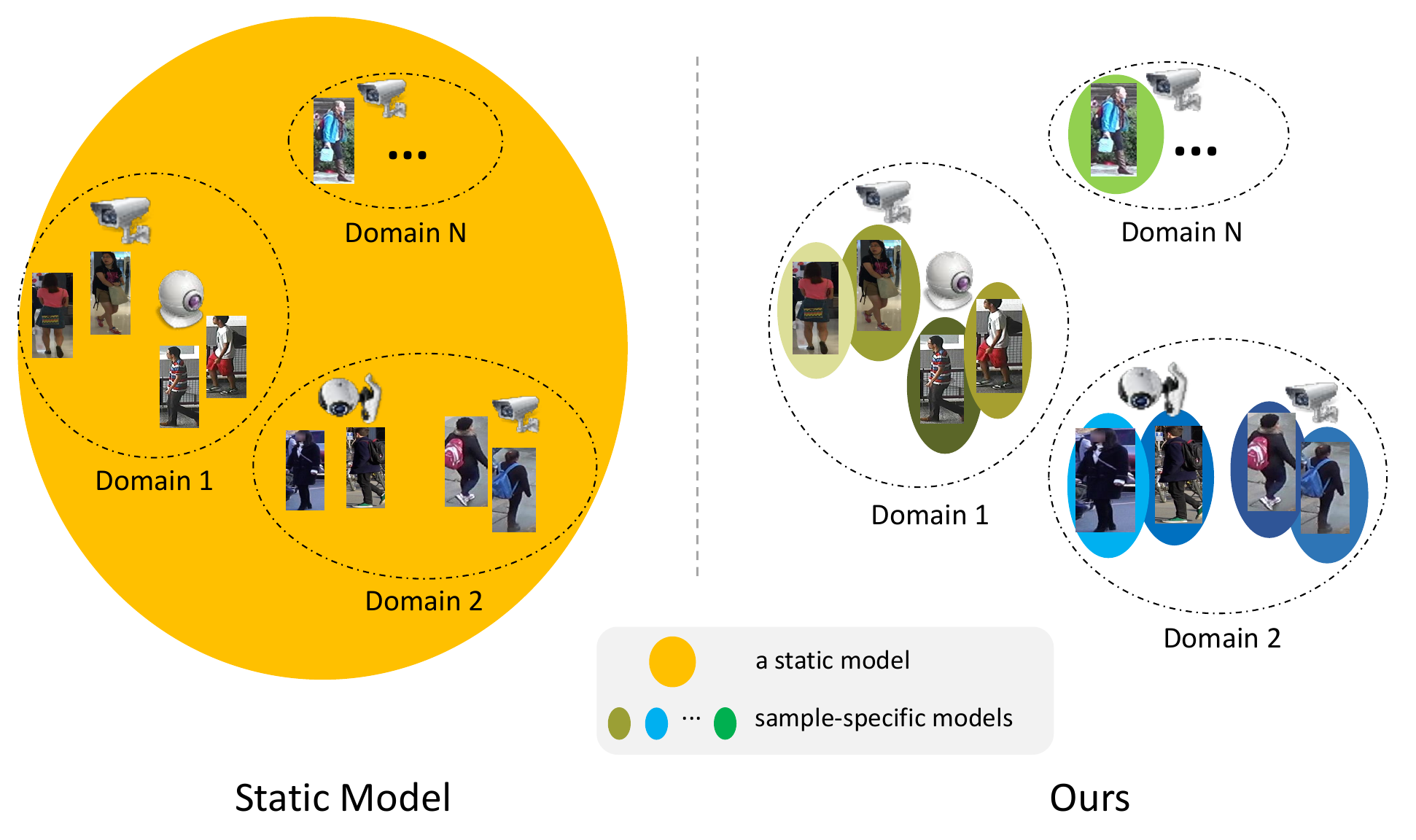}
}
\caption{
Data in different domains usually have large data distribution shifts, it will be difficult for a traditional static model to deal with the data of all domains at the same time. 
The static network needs to learn a very powerful feature extraction ability to handle these datasets, it brings a heavy learning burden to the network.
Unlike the traditional fixed network, DCSD uses internal domain-related factor, such as features from the sample, and external domain-related factors, such as dataset IDs and camera IDs, to dynamically generate the sample-specific network.
It can reduce the learning burden of the model and makes the model easier to learn. Each colored ellipse represents a model, best viewed in color.
}
\label{fig:Figure1}
\end{center}
\end{figure}

In this work, we propose a Domain-Camera-Sample Dynamic network (DCSD) to solve the domain conflict problem.
When there is little training data in the current domain, DCSD can combine the data of other domains for joint training without domain conflict problems, so as to improve the performance of the model when deployed in the current domain.
Unlike the fixed network parameters, DCSD can simultaneously consider the domain, camera, and sample information to generate the sample-specific network. 
We use three different levels of domain-related information to generate network parameters, the key insight is that datasets, cameras, and samples are different levels of factors that lead to shifts in data distribution.
For example, there are location and season bias among different datasets, illumination and angle bias among different cameras, and micro-specific bias among samples. 

In summary, the contribution of this study is three-fold:
\begin{itemize}
\item 
In this work, we deliver some surprising findings that the performance of joint training in multiple-person ReID datasets is not as effective as that of individual training in each domain.

\item 
We propose a Domain-Camera-Sample Dynamic network (DCSD) to solve the domain conflict problem. It simultaneously considering the domain, camera, and sample information to generate the sample-specific network. 

\item 
By conducting extensive experiments on various person ReID datasets, we demonstrate the superior performance of the proposed DCSD and show that it can lead to consistent performance improvement when the model is jointly trained on multiple domains. And we also achieve new state-of-the-art performance on multiple ReID benchmarks.

\end{itemize}

\section{Related Work}
\subsection{Domain Adaptation}
Domain gap often exists between multiple datasets, and many researchers study on the domain adaptation problem.
~\cite{ganin2015unsupervised} propose unsupervised domain adaptation of deep feed-forward architectures, which allows large-scale training based on a large amount of annotated data in the source domain and a large amount of unannotated data in the target domain. Similar to many previous shallow and deep domain adaptive techniques, the adaptation is achieved through aligning the distributions of features across the two domains.
Such ideas have appeared in other works~\cite{2016Return,sun2016deep}.
Recent GAN methods~\cite{bousmalis2017unsupervised,liu2016coupled,hoffman2018cycada} use an adversarial approach to learn a transformation in the pixel space from one domain to another. The methods seek to find a domain-invariant feature space using the maximum mean discrepancy for this purpose is also achieved superior results to a certain extent. 
Dynamic transfer for multi-source domain adaptation ~\cite{li2021dynamic} across multiple domains and unifies multiple source domains into a single source domain, which simplifies the alignment between source and target domains. 
~\cite{Bai_2021_CVPR} proposed a multi-source framework from two perspectives,
i.e. domain-specific view and domain-fusion view, for the unsupervised domain adaptive person ReID.
The above domain adaptation methods focus on how to improve the performance of the target domain, and the target domain is different from the source domain. But the target domain studied in this paper is one of source domains.

\subsection{Dynamic Convolution}
The convolution operator is the core of convolutional neural networks (CNNs) and occupies the most computation cost. In order to address the issue that light-weight convolutional neural networks (CNNs) suffer performance degradation as their low computational budgets constrain both the depth (number of convolution layers) and the width (number of channels) of CNNs, dynamic convolution, a new design method that increases model complexity without increasing the network depth or width. 

Instead of using a single convolution kernel per layer, \cite{2020Dynamic} aggregates multiple parallel convolution kernels dynamically based upon their attention, which are input dependent. Assembling multiple kernels is not only computationally efficient due to the small kernel size, but also has more representation power since these kernels are aggregated in a non-linear way via attention. \cite{zhang2020dynet} propose a novel dynamic convolution method to adaptively generate convolution kernels based on image content, which reduces the redundant computation cost existed in conventional convolution kernels. \cite{Li_2021_CVPR} propose an effective and efficient operator for visual representation learning, reversing the design principles of convolution and generalizing the formulation of self-attention, which are able to disclose the underlying relationship between self-attention and convolution. \cite{Zhou_2021_CVPR} propose a lightweight content-adaptive filtering technique called DDF, which the main subject is to predict decoupled spatial and channel dynamic filters. This method can seamlessly replace standard convolution layers, consistently improving the performance of ResNets while also reducing model parameters and computational costs. 

Most of these methods use dynamic convolution to get a larger model capacity. These methods generate different parameters according to different input features.
And domain, camera, and sample are the three factors that cause large domain shifts in person ReID.
The proposed DCSD method uses three information to generate model parameters, It reduces the domain conflict between multiple datasets and improves the performance of multi-domain joint training.

\begin{figure}[t]
\begin{center}
\resizebox{0.5\textwidth}{!}{
\includegraphics[width=\linewidth]{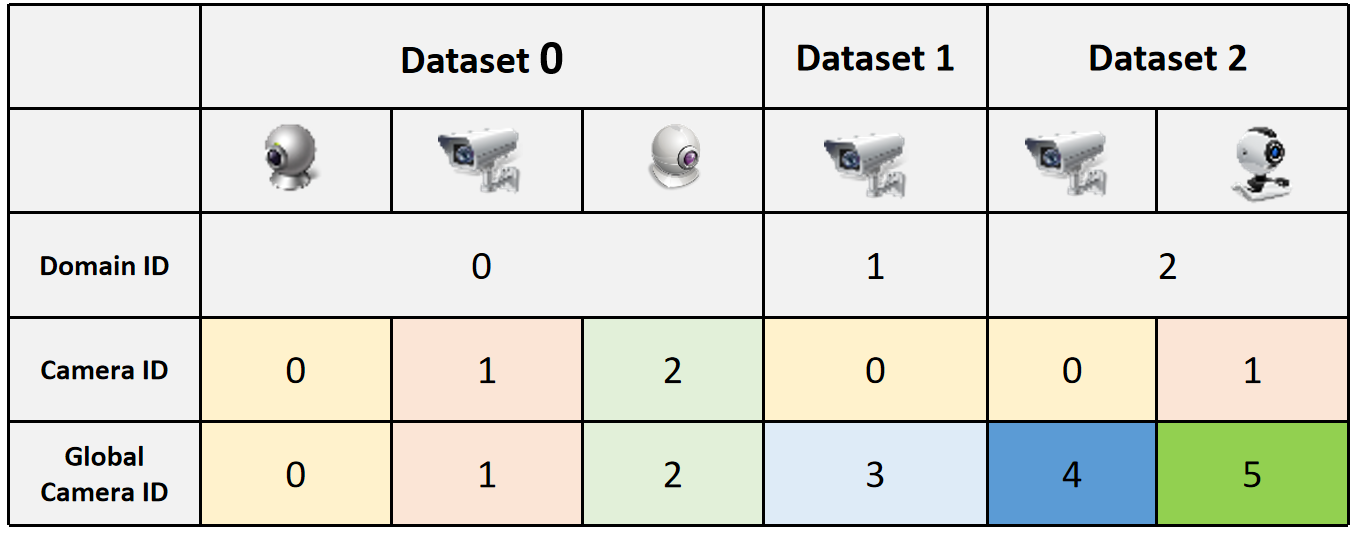}
}
\caption{
Example of domainID, camera ID and global camera ID.
Each dataset is treated as a domain, so each dataset has a unique domain ID.
Since camera ID already existed in the datasets, there may be the same camera ID among datasets, so we use the global camera ID in our experiments by default. Best viewed in color.
}
\label{fig:global_camera}
\end{center}
\end{figure}

\begin{figure*}[t]
\begin{center}
\resizebox{1.0\textwidth}{!}{
\includegraphics[width=\linewidth]{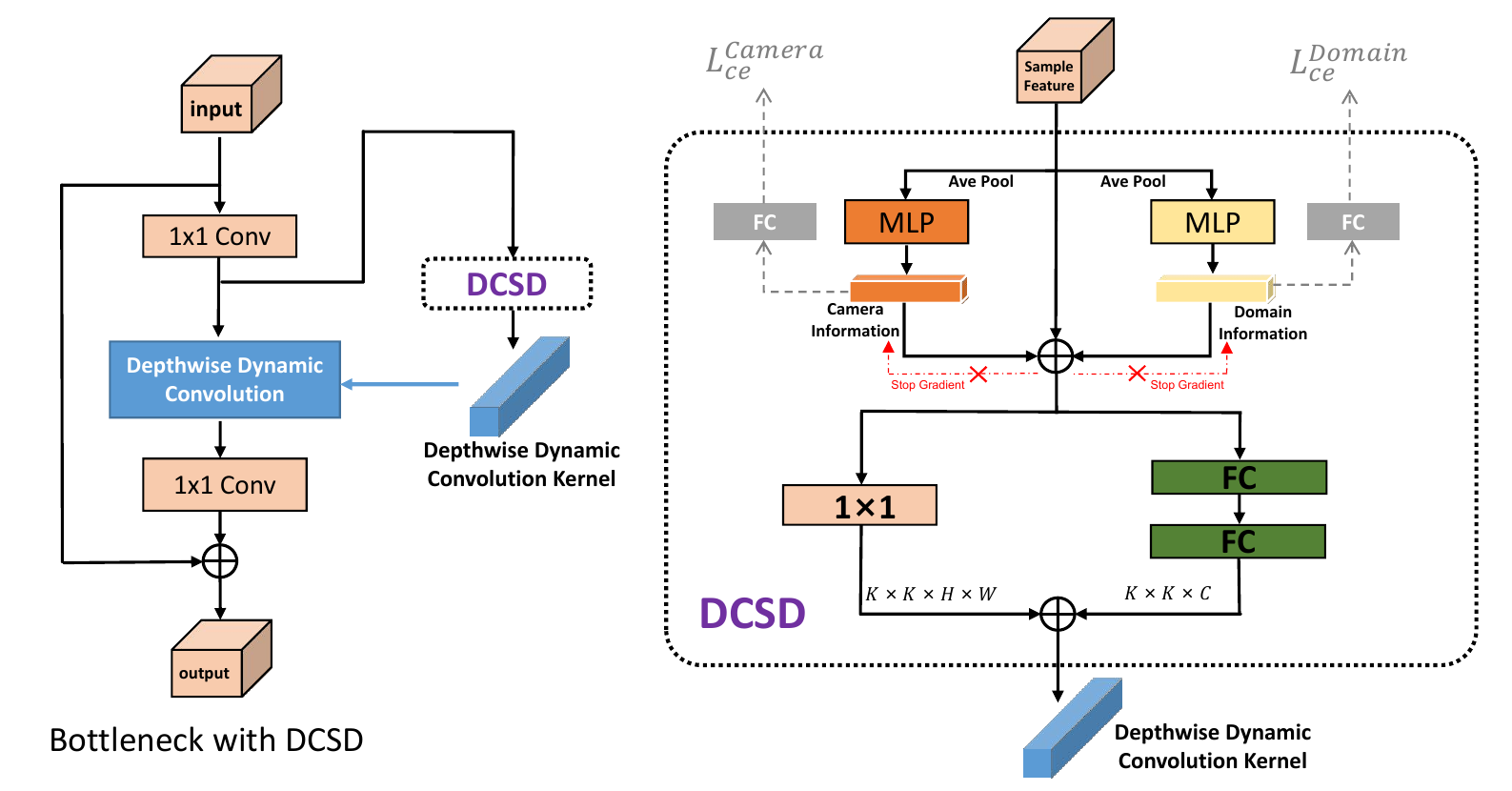}
}
\caption{
The architecture of DCSD module. 
The DCSD is a $K \times K$ depthwise dynamic convolution, and its parameters are
generated by its domain, camera and sample information.
The domain information and camera information are predicted by the input feature with two MLPs, and then the input feature, domain information and camera information are added together to generating the convolution parameters. The MLP is consists of two fully connection layers, and hidden dimension is $1/4$ of the input dimension. In the process of generating convolution parameters, we use two branches to reduce memory consumption. Best viewed in color.
}
\label{fig:architecture}
\end{center}
\end{figure*}

\section{Approach}
In person Re-ID, there is a common problem of domain gap between different datasets, and even sub-domain gap exist between different cameras in the same dataset. Therefore, there are a lot of works on domain transfer for person Re-ID, that is, training in one domain and testing in another domain. 
However, few researchers have studied the domain conflict problem when  training in multiple domains and testing in one of them. 

From our experiments, we found that the performance of multiple Re-ID datasets in joint training is even worse than that of trained on single dataset. Further more, datasets with many cameras may have sub-domain conflicts problems.
A static model is difficult to handle multiple datasets with large domain variances. Therefore, we propose a Domain-Camera-Sample Dynamic convolution module, which can generate a sample specific network according to the domain and camera information of each sample, so as to reduce the learning burden of the network and improve performance.

As shown in Figure ~\ref{fig:architecture}, 
the DCSD is a $K \times K$ depthwise dynamic convolution, and its parameters are generated by its domain, camera and sample information.
The domain information and camera information are predicted by the input feature with two MLPs, and then the input feature, domain information and camera information are added together to generating the convolution parameters. 
The input feature (sample information) is from the sample, which is an internal factor, while the domain information and camera information are external factors.
In the process of generating convolution parameters, we use two branches, spatial branch and channel branch, to reduce the memory consumption.
The spatial branch generates the features ($K\times K\times H\times W$) through a $1\times 1$ convolution, and the channel branch generates the features ($K\times K\times C$) through two layers of fully connection. During the convolution operation, the corresponding ($K\times K$) kernel is obtained from the outputs from the two branches according to the spatial position and channel position, and they are multiplied to obtain the final ($K\times K$) kernel for the current position.
We can use DCSD module to replace the standard convolution in the classical model to realize the DCSD model. In this paper, we use DCSD ($K=3$) to replace all ($3\times 3$) convolution layers in ResNet50, which is called ResNet50-DCSD.

\subsection{DCSD Module}
To better understand our approach, we give a brief review of the standard convolution. The input feature can be written as
$X \in \mathbb{R}^{H\times W\times C}$, where $H$, $W$ and $C$ are height, width and number of channels of the input $X$ respectively. 
The convolution kernel of the standard convolution is $W \in \mathbb{R}^{K\times K\times C\times C' }$, where $K$ is the kernel size. If $pad = K-1$ and $stride = 1$, then the output feature is $output \in \mathbb{R}^{H\times W\times C'}$, $C'$ is the output channels. And the convolution operation (ignore the bias) can be written as 
\begin{equation}
   \mathbf{Output}_{k,l,c'} = \sum_{i,j,c} \mathbf{W}_{i,j,c,c'} \cdot \mathbf{X}_{k+i-1,l+j-1,c}
  \label{eq:stdconv}
\end{equation}

In dynamic convolution, the model needs to generate convolution parameters before the convolution operation. If the convolution kernel parameters are generated directly for standard convolution, the model size and computational cost will be extremely vast. In order to decrease the model size and computation cost, some methods~\cite{Li_2021_CVPR,Zhou_2021_CVPR} use depthwise dynamic convolution instead of standard convolution. We also use depthwise dynamic convolution in DCSD, but there are some subtle details in DCSD are distinct from the normal methodologies. The parameters of our depthwise version of dynamic convolution are jointly generated by domain, camera and sample information. The DCSD module can be written as 
\begin{equation}
\label{eq:dyconv}
\resizebox{0.42\textwidth}{!}{
\begin{math}
\begin{aligned}
  \hat{\mathbf{W}} = DCSD(\mathbf{X}, &\mathbf{Domain}.detach(), \mathbf{Camera}.detach()),\\
  \hat{\mathbf{Output}_{k,l,c}} &= \sum_{i,j} \hat{\mathbf{W}}_{i,j,c} \cdot \mathbf{X}_{k+i-1,l+j-1,c},
\end{aligned}
\end{math}
}
\end{equation}
where $X$ is the feature extracted from the sample, the information of $Domain$ and $Camera$ is predicted by $X$ through two MLPs, and the domain ID and camera ID are used for supervision training by cross entropy losses. 
We cut off the gradient propagation for $Domain$ and $Camera$ for making $Domain$ and $Camera$ only supervised by domain ID and camera ID.
$detach()$ means that it detaches the output from the computational graph, so no gradient will be back propagated along the variable. As Figure~\ref{fig:architecture} shows, DCSD generates convolution parameters through domain, camera and sample information, and then the convolution parameters are applied to the input feature.

The models in our experiments are trained by jointly minimizing the triplet loss and the cross-entropy losses. The cross-entropy losses include sample ID classification loss, domain ID classification losses and camera ID classification losses.
Specifically, for $N$ vehicle image samples selected from $M$ IDs, the triplet loss can be formulated as 
\begin{equation}
\begin{aligned}
    L_{tri}=\frac{1} {N}\sum^{N}_{i=1} \big[ \, D(\mathbf{x_i}, \mathbf{x_i}^p) - D(\mathbf{x_i}, \mathbf{x_i}^n) + \alpha  \, \big]_+,
\end{aligned}
\end{equation}
where $\mathbf{x_i}$, $\mathbf{x_i}^p$, and $\mathbf{x_i}^n$ 
denote the anchor, positive and negative samples respectively. In practice, we often choose the farthest positive sample and the closest negative sample in the batch to form a hard triplet. $D(\cdot)$ is a distance metric and $\alpha$ is a pre-defined margin scalar (e.g., $0.3$). The $[\cdot]_{+}$ denotes $max([\cdot], 0)$. 

The cross-entropy loss can be defined as
\begin{equation}\label{Fun:CE}
L_{ce}=-\frac{1} {N}\sum^{N}_{i=1}\sum^{M}_{j=1}[y_i=j] \cdot \mathrm{log}(Prob_{i,j}),
\end{equation}
where $[\cdot]$ denotes the indicator function and $y_{i}$ is the ground truth 
ID for the $i$-th sample.
$Prob_{i,j}$ is the predicted probability for $i$-th sample belonging to the ID 
$j$. Note that, in our experiment, there are three different types of IDs: domain ID, camera ID and sample ID. 
Then the total loss is:
\begin{equation}
\label{Fun:total_loss}
\resizebox{0.43\textwidth}{!}{
\begin{math}
\begin{aligned}
L_{total} = L_{tri} + L_{ce}^{Sample} + \frac{1}{B}\sum^{B}_{i=1}L_{ce}^{Domain} + \frac{1}{B}\sum^{B }_{i=1}L_{ce}^{Camera}
\end{aligned}
\end{math}
}
\end{equation}
where $B$ is the number of BottleNeck in the model.

\subsection{Global Camera ID}
Our DCSD requires domain ID and camera ID as input information. We number datasets and use the index as their domain IDs, so each dataset has a unique domain ID. 
Because camera ID already existed in the datasets, there may be the same camera ID among datasets, so we need to use the global camera ID. In other words, we uniformly number all cameras among multiple datasets. The numbering method is shown in figure ~\ref{fig:global_camera}.

\begin{table*}[th]
	\caption{Performance (\%) comparisons with the state-of-the-arts on CUHK03, Market1501, Duke and MSMT17. ``\dag'' means by our implementation. \textbf{Bold} and \emph{Italic} fonts represent the best and second best respectively. The ``Joint Training Gain'' is the performance gap between ``BoT/AGW (Joint Training)'' and ``BoT/AGW + DCSD (Joint Training)''.}
	\label{tab:SOTA_person}
		\centering
		\footnotesize
	\scalebox{1.0}{
    \begin{tabular}{crrrrrrrrrrr}
    	\hline \hline
    	\multicolumn{1}{c}{\multirow{2}{*}{Method}} & \multicolumn{2}{c}{CUHK03(L)} & \multicolumn{2}{c}{CUHK-SYSU} & \multicolumn{2}{c}{Market1501} & \multicolumn{2}{c}{Duke} & \multicolumn{2}{c}{MSMT17} \bigstrut[t]\\

    	\cline{2-11} \multicolumn{1}{c}{} & \multicolumn{1}{c}{Rank-1} & \multicolumn{1}{c}{mAP} & \multicolumn{1}{c}{Rank-1} & \multicolumn{1}{c}{mAP} & \multicolumn{1}{c}{Rank-1} & \multicolumn{1}{c}{mAP} & \multicolumn{1}{c}{Rank-1} & \multicolumn{1}{c}{mAP} & \multicolumn{1}{c}{Rank-1} & \multicolumn{1}{c}{mAP} \bigstrut\\
    	\hline
    	 
    	\multicolumn{1}{l}{IDE~\cite{sun2017beyond}} & \multicolumn{1}{c}{43.8 } & \multicolumn{1}{c}{38.9 } & \multicolumn{1}{c}{-} & \multicolumn{1}{c}{-} & \multicolumn{1}{c}{85.3 } & \multicolumn{1}{c}{68.5 } & \multicolumn{1}{c}{73.2 } & \multicolumn{1}{c}{52.8 } & \multicolumn{1}{c}{-} & \multicolumn{1}{c}{- } \bigstrut[t]\\ 
    	
    	\multicolumn{1}{l}{Gp-reid~\cite{almazan2018re}} & \multicolumn{1}{c}{-} & \multicolumn{1}{c}{-} & \multicolumn{1}{c}{-} & \multicolumn{1}{c}{-} & \multicolumn{1}{c}{92.2 } & \multicolumn{1}{c}{81.2 } & \multicolumn{1}{c}{85.2 } & \multicolumn{1}{c}{72.8 } & \multicolumn{1}{c}{-} & \multicolumn{1}{c}{- }\bigstrut[t]\\
    	 
    	\multicolumn{1}{l}{MGCAM~\cite{song2018mask}} & \multicolumn{1}{c}{50.1 } & \multicolumn{1}{c}{50.2 } & \multicolumn{1}{c}{- } & \multicolumn{1}{c}{- } & \multicolumn{1}{c}{83.8 } & \multicolumn{1}{c}{74.3 } & \multicolumn{1}{c}{-} & \multicolumn{1}{c}{-}& \multicolumn{1}{c}{-} & \multicolumn{1}{c}{- } \bigstrut[t]\\
    	
    	\multicolumn{1}{l}{MaskReID~\cite{qi2018maskreid}} &  \multicolumn{1}{c}{-}  &  \multicolumn{1}{c}{-}  &  \multicolumn{1}{c}{-}  &  \multicolumn{1}{c}{-}  & \multicolumn{1}{c}{90.0 } & \multicolumn{1}{c}{70.3 } & \multicolumn{1}{c}{78.9 } & \multicolumn{1}{c}{61.9 }& \multicolumn{1}{c}{-} & \multicolumn{1}{c}{- } \bigstrut[t]\\
    	
    	\multicolumn{1}{l}{AACN~\cite{xu2018attention}} & \multicolumn{1}{c}{-} & \multicolumn{1}{c}{-} & \multicolumn{1}{c}{-} & \multicolumn{1}{c}{-} & \multicolumn{1}{c}{85.9 } & \multicolumn{1}{c}{66.9 } & \multicolumn{1}{c}{76.8 } & \multicolumn{1}{c}{59.3 } & \multicolumn{1}{c}{-} & \multicolumn{1}{c}{- }\bigstrut[t]\\
    	
    	\multicolumn{1}{l}{SPReID~\cite{kalayeh2018human}} & \multicolumn{1}{c}{-} & \multicolumn{1}{c}{-} & \multicolumn{1}{c}{-} & \multicolumn{1}{c}{-} & \multicolumn{1}{c}{92.5 } & \multicolumn{1}{c}{81.3 } & \multicolumn{1}{c}{84.4 } & \multicolumn{1}{c}{71.0 }  & \multicolumn{1}{c}{-} & \multicolumn{1}{c}{- }\bigstrut[t]\\
    	    
    	\multicolumn{1}{l}{HA-CNN~\cite{li2018harmonious}} & \multicolumn{1}{c}{44.4 } & \multicolumn{1}{c}{41.0 } & \multicolumn{1}{c}{- } & \multicolumn{1}{c}{- } & \multicolumn{1}{c}{91.2 } & \multicolumn{1}{c}{75.7 } & \multicolumn{1}{c}{80.5 } & \multicolumn{1}{c}{63.8 } & \multicolumn{1}{c}{-} & \multicolumn{1}{c}{- }\bigstrut[t]\\
    	
    	\multicolumn{1}{l}{DuATM~\cite{si2018dual}} & \multicolumn{1}{c}{-} & \multicolumn{1}{c}{-} & \multicolumn{1}{c}{-} & \multicolumn{1}{c}{-} & \multicolumn{1}{c}{91.4 } & \multicolumn{1}{c}{76.6 } & \multicolumn{1}{c}{81.8 } & \multicolumn{1}{c}{64.6 }  & \multicolumn{1}{c}{-} & \multicolumn{1}{c}{- }\bigstrut[t]\\
    	
    	\multicolumn{1}{l}{Mancs~\cite{wang2018mancs}} & \multicolumn{1}{c}{69.0 } & \multicolumn{1}{c}{63.9 } & \multicolumn{1}{c}{- } & \multicolumn{1}{c}{- } & \multicolumn{1}{c}{93.1 } & \multicolumn{1}{c}{82.3 } & \multicolumn{1}{c}{84.9 } & \multicolumn{1}{c}{71.8 } & \multicolumn{1}{c}{-} & \multicolumn{1}{c}{- }\bigstrut[t]\\
    	
    	\multicolumn{1}{l}{MGN~\cite{wang2018learning}} & \multicolumn{1}{c}{68.0 } & \multicolumn{1}{c}{67.4 } & \multicolumn{1}{c}{- } & \multicolumn{1}{c}{- } & \multicolumn{1}{c}{95.7 } & \multicolumn{1}{c}{86.9 } & \multicolumn{1}{c}{88.7 } & \multicolumn{1}{c}{78.4 } & \multicolumn{1}{c}{- } & \multicolumn{1}{c}{- } \bigstrut[t]\\
    	
    	\multicolumn{1}{l}{HPM~\cite{fu2019horizontal}} & \multicolumn{1}{c}{63.9 } & \multicolumn{1}{c}{57.5 } & \multicolumn{1}{c}{-} & \multicolumn{1}{c}{-} & \multicolumn{1}{c}{94.2 } & \multicolumn{1}{c}{82.7 } & \multicolumn{1}{c}{86.6 } & \multicolumn{1}{c}{74.3 }& \multicolumn{1}{c}{-} & \multicolumn{1}{c}{- } \bigstrut[t]\\
    	
    	\multicolumn{1}{l}{DSA-reID~\cite{zhang2019densely}} & \multicolumn{1}{c}{78.9 } & \multicolumn{1}{c}{75.2 } & \multicolumn{1}{c}{-} & \multicolumn{1}{c}{- } & \multicolumn{1}{c}{95.7} & \multicolumn{1}{c}{87.6 } & \multicolumn{1}{c}{86.2 } & \multicolumn{1}{c}{74.3 }& \multicolumn{1}{c}{-} & \multicolumn{1}{c}{- } \bigstrut[t]\\
    	
    	\multicolumn{1}{l}{OSNet~\cite{2019Omni}} & \multicolumn{1}{c}{72.3} & \multicolumn{1}{c}{67.8} & \multicolumn{1}{c}{-} & \multicolumn{1}{c}{-} & \multicolumn{1}{c}{94.8 } & \multicolumn{1}{c}{84.9 } & \multicolumn{1}{c}{88.6 } & \multicolumn{1}{c}{73.5 } & \multicolumn{1}{c}{78.7 } & \multicolumn{1}{c}{52.9 } \bigstrut[t]\\

    	\multicolumn{1}{l}{IANet~\cite{2020IANet}} & \multicolumn{1}{c}{72.4} & \multicolumn{1}{c}{- } & \multicolumn{1}{c}{- } & \multicolumn{1}{c}{- } & \multicolumn{1}{c}{94.4 } & \multicolumn{1}{c}{83.1 } & \multicolumn{1}{c}{87.1} & \multicolumn{1}{c}{73.4} & \multicolumn{1}{c}{75.5 } & \multicolumn{1}{c}{46.8} \bigstrut[t]\\
    	
        \multicolumn{1}{l}{HOReid~\cite{wang2020high}} & \multicolumn{1}{c}{-} & \multicolumn{1}{c}{-} & \multicolumn{1}{c}{-} & \multicolumn{1}{c}{-} & \multicolumn{1}{c}{94.2 } & \multicolumn{1}{c}{84.9} & \multicolumn{1}{c}{86.9} & \multicolumn{1}{c}{75.6} & \multicolumn{1}{c}{-} & \multicolumn{1}{c}{- } \bigstrut[t]\\
        
        \multicolumn{1}{l}{ISP~\cite{zhu2020identity}} & \multicolumn{1}{c}{76.5} & \multicolumn{1}{c}{74.1} & \multicolumn{1}{c}{-} & \multicolumn{1}{c}{-} & \multicolumn{1}{c}{95.3 } & \multicolumn{1}{c}{88.6} & \multicolumn{1}{c}{\emph{89.6}} & \multicolumn{1}{c}{80.0} & \multicolumn{1}{c}{-} & \multicolumn{1}{c}{- } \bigstrut[t]\\
        
    	
    	\multicolumn{1}{l}{TransReID (DeiT-S/16)~\cite{he2021transreid}} & \multicolumn{1}{c}{-} & \multicolumn{1}{c}{-} & \multicolumn{1}{c}{-} & \multicolumn{1}{c}{-} & \multicolumn{1}{c}{-} & \multicolumn{1}{c}{-} & \multicolumn{1}{c}{-} & \multicolumn{1}{c}{-} & \multicolumn{1}{c}{76.3} & \multicolumn{1}{c}{55.2}  \bigstrut[t]\\   
    	
    	\hline   	
    	\multicolumn{1}{l}{BoT$^{\dag}$~\cite{luo2019bag}} & \multicolumn{1}{c}{69.1} & \multicolumn{1}{c}{67.3} & \multicolumn{1}{c}{87.2} & \multicolumn{1}{c}{86.0} & \multicolumn{1}{c}{94.2} & \multicolumn{1}{c}{86.1} & \multicolumn{1}{c}{86.4} & \multicolumn{1}{c}{76.8} & \multicolumn{1}{c}{74.1} & \multicolumn{1}{c}{50.2}  \bigstrut[t]\\
    	\multicolumn{1}{l}{BoT (Joint Training)} & \multicolumn{1}{c}{67.5} & \multicolumn{1}{c}{65.7} & \multicolumn{1}{c}{90.5} & \multicolumn{1}{c}{89.3} & \multicolumn{1}{c}{93.7} & \multicolumn{1}{c}{84.6} & \multicolumn{1}{c}{86.0} & \multicolumn{1}{c}{75.9} & \multicolumn{1}{c}{72.9} & \multicolumn{1}{c}{49.5}  \bigstrut[t]\\
    	\multicolumn{1}{l}{BoT-DCSD} & \multicolumn{1}{c}{66.8} & \multicolumn{1}{c}{66.5} & \multicolumn{1}{c}{87.2} & \multicolumn{1}{c}{85.6} & \multicolumn{1}{c}{95.3} & \multicolumn{1}{c}{87.8} & \multicolumn{1}{c}{88.6} & \multicolumn{1}{c}{79.0} & \multicolumn{1}{c}{79.8} & \multicolumn{1}{c}{58.2}  \bigstrut[t]\\
    	\multicolumn{1}{l}{BoT-DCSD (Joint Training)} & \multicolumn{1}{c}{\emph{79.4}} & \multicolumn{1}{c}{\emph{78.0}} & \multicolumn{1}{c}{\emph{92.4}} & \multicolumn{1}{c}{\emph{91.3}} & \multicolumn{1}{c}{\textbf{95.9}} & \multicolumn{1}{c}{\textbf{89.5}} & \multicolumn{1}{c}{\emph{89.6}} & \multicolumn{1}{c}{\emph{80.6}} & \multicolumn{1}{c}{\emph{80.3}} & \multicolumn{1}{c}{\emph{61.5}}  \bigstrut[t]\\
    	
    	\multicolumn{1}{l}{Joint Training Gain} & \multicolumn{1}{c}{+11.9} & \multicolumn{1}{c}{+12.3} & \multicolumn{1}{c}{+1.9} & \multicolumn{1}{c}{+2.0} & \multicolumn{1}{c}{+2.2} & \multicolumn{1}{c}{+4.9} & \multicolumn{1}{c}{+3.6} & \multicolumn{1}{c}{+4.7} & \multicolumn{1}{c}{+7.4} & \multicolumn{1}{c}{+12.0}  \bigstrut[t]\\
    	
    	\hline
    	\multicolumn{1}{l}{AGW$^{\dag}$~\cite{2021AGW}} & \multicolumn{1}{c}{75.1} & \multicolumn{1}{c}{73.4} & \multicolumn{1}{c}{90.1} & \multicolumn{1}{c}{88.7} & \multicolumn{1}{c}{95.4} & \multicolumn{1}{c}{88.5} & \multicolumn{1}{c}{89.2} & \multicolumn{1}{c}{79.4} & \multicolumn{1}{c}{78.0} & \multicolumn{1}{c}{54.3}  \bigstrut[t]\\    
    	
    	\multicolumn{1}{l}{AGW (Joint Training)} & \multicolumn{1}{c}{70.3} & \multicolumn{1}{c}{69.1} & \multicolumn{1}{c}{91.8} & \multicolumn{1}{c}{90.8} & \multicolumn{1}{c}{94.0} & \multicolumn{1}{c}{86.2} & \multicolumn{1}{c}{88.0} & \multicolumn{1}{c}{78.5} & \multicolumn{1}{c}{75.5} & \multicolumn{1}{c}{52.6}  \bigstrut[t]\\
    	
    	\multicolumn{1}{l}{AGW-DCSD} & \multicolumn{1}{c}{77.2} & \multicolumn{1}{c}{74.3} & \multicolumn{1}{c}{84.9} & \multicolumn{1}{c}{83.6} & \multicolumn{1}{c}{95.3} & \multicolumn{1}{c}{88.1}  & \multicolumn{1}{c}{\emph{89.6}} & \multicolumn{1}{c}{80.3} & \multicolumn{1}{c}{80.1} & \multicolumn{1}{c}{60.7}  \bigstrut[t]\\
    	
    	\multicolumn{1}{l}{AGW-DCSD (Joint Training)} & \multicolumn{1}{c}{\textbf {80.9}} & \multicolumn{1}{c}{\textbf{78.8}} & \multicolumn{1}{c}{\textbf{92.7}} & \multicolumn{1}{c}{\textbf{91.4}} & \multicolumn{1}{c}{\emph{95.7}} & \multicolumn{1}{c}{\emph{89.3}} & \multicolumn{1}{c}{\textbf{90.0}} & \multicolumn{1}{c}{\textbf{81.2}} & \multicolumn{1}{c}{\textbf{81.8}} & \multicolumn{1}{c}{\textbf{62.9}}  \bigstrut[t]\\
    	
    	\multicolumn{1}{l}{Joint Training Gain} & \multicolumn{1}{c}{+10.6} & \multicolumn{1}{c}{+9.7} & \multicolumn{1}{c}{+0.9} & \multicolumn{1}{c}{+0.6} & \multicolumn{1}{c}{+1.7} & \multicolumn{1}{c}{+3.1} & \multicolumn{1}{c}{+2.0} & \multicolumn{1}{c}{+2.7} & \multicolumn{1}{c}{+6.3} & \multicolumn{1}{c}{+10.3}  \bigstrut[t]\\
    	
    	\hline \hline
    \end{tabular}
    }
\end{table*}

\begin{table*}[th]
	\caption{The multi-domain joint training performance (\%) comparisons with the state-of-the-arts domain generalization methods on CUHK03, Market1501, Duke and MSMT17. Each row is trained on the same source domain, and then test the performance on different target domains.~\textbf{Bold} fonts represent the best, ``*'' means by our implementation.}
	\label{tab:DG}
		\centering
		\footnotesize
    \begin{tabular}{ccrrrrrrrrrrr}
    	\hline \hline
    	\multicolumn{1}{c}{\multirow{2}{*}{Method}} & \multicolumn{1}{c}{\multirow{2}{*}{Source}} & \multicolumn{2}{c}{CUHK03(L)} & \multicolumn{2}{c}{CUHK-SYSU} & \multicolumn{2}{c}{Market1501} & \multicolumn{2}{c}{Duke} & \multicolumn{2}{c}{MSMT17} \bigstrut[t]\\
    	\cline{3-12} \multicolumn{2}{c}{} & \multicolumn{1}{c}{Rank-1} & \multicolumn{1}{c}{mAP} & \multicolumn{1}{c}{Rank-1} & \multicolumn{1}{c}{mAP} & \multicolumn{1}{c}{Rank-1} & \multicolumn{1}{c}{mAP} & \multicolumn{1}{c}{Rank-1} & \multicolumn{1}{c}{mAP} & \multicolumn{1}{c}{Rank-1} & \multicolumn{1}{c}{mAP} \bigstrut\\
    	\hline

    	\multicolumn{1}{l}{\multirow{6}{*}{BoT (Baseline) ~\cite{luo2019bag}}} & CUHK03(L) &\multicolumn{1}{c}{69.1} & \multicolumn{1}{c}{67.3} & \multicolumn{1}{c}{70.2} & \multicolumn{1}{c}{67.3} & \multicolumn{1}{c}{46.3} & \multicolumn{1}{c}{22.3} & \multicolumn{1}{c}{19.9} & \multicolumn{1}{c}{10.5} & \multicolumn{1}{c}{7.1} & \multicolumn{1}{c}{2.1}  \bigstrut[t]\\
     	\cline{3-12} & CUHK-SYSU &\multicolumn{1}{c}{3.9} & \multicolumn{1}{c}{4.4} & \multicolumn{1}{c}{87.2} & \multicolumn{1}{c}{86.0} & \multicolumn{1}{c}{46.2} & \multicolumn{1}{c}{23.5} & \multicolumn{1}{c}{24.5} & \multicolumn{1}{c}{14.5} & \multicolumn{1}{c}{12.8} & \multicolumn{1}{c}{4.8}  \bigstrut[t]\\
     	\cline{3-12} & Market1501 &\multicolumn{1}{c}{4.0} & \multicolumn{1}{c}{4.4} & \multicolumn{1}{c}{74.4} & \multicolumn{1}{c}{70.8} & \multicolumn{1}{c}{94.2} & \multicolumn{1}{c}{86.1} & \multicolumn{1}{c}{27.6} & \multicolumn{1}{c}{14.8} & \multicolumn{1}{c}{7.7} & \multicolumn{1}{c}{2.4}  \bigstrut[t]\\
     	\cline{3-12} & Duke &\multicolumn{1}{c}{4.8} & \multicolumn{1}{c}{5.2} & \multicolumn{1}{c}{66.1} & \multicolumn{1}{c}{62.0} & \multicolumn{1}{c}{48.0} & \multicolumn{1}{c}{22.2} & \multicolumn{1}{c}{86.4} & \multicolumn{1}{c}{76.8} & \multicolumn{1}{c}{12.8} & \multicolumn{1}{c}{3.9}  \bigstrut[t]\\
     	\cline{3-12} & MSMT17 &\multicolumn{1}{c}{9.7} & \multicolumn{1}{c}{10.3} & \multicolumn{1}{c}{79.5} & \multicolumn{1}{c}{76.7} & \multicolumn{1}{c}{56.4} & \multicolumn{1}{c}{28.7} & \multicolumn{1}{c}{49.6} & \multicolumn{1}{c}{33.5} & \multicolumn{1}{c}{74.1} & \multicolumn{1}{c}{50.2}  \bigstrut[t]\\
     	\cline{3-12} & Joint Training &\multicolumn{1}{c}{67.5} & \multicolumn{1}{c}{65.7} & \multicolumn{1}{c}{90.5} & \multicolumn{1}{c}{89.3} & \multicolumn{1}{c}{93.7} & \multicolumn{1}{c}{84.6} & \multicolumn{1}{c}{86.0} & \multicolumn{1}{c}{75.9} & \multicolumn{1}{c}{72.9} & \multicolumn{1}{c}{49.5}  \bigstrut[t]\\
     	
    	\hline
     	\multicolumn{1}{l}{\multirow{6}{*}{BoT-DN*~\cite{dual_norm_bmvc2019}}} & CUHK03(L) &\multicolumn{1}{c}{59.6} & \multicolumn{1}{c}{57.9} & \multicolumn{1}{c}{68.6} & \multicolumn{1}{c}{65.2} & \multicolumn{1}{c}{58.1} & \multicolumn{1}{c}{30.6} & \multicolumn{1}{c}{43.1} & \multicolumn{1}{c}{24.6} & \multicolumn{1}{c}{18.3} & \multicolumn{1}{c}{5.7}  \bigstrut[t]\\
     	\cline{3-12} & CUHK-SYSU &\multicolumn{1}{c}{5.1} & \multicolumn{1}{c}{6.5} & \multicolumn{1}{c}{79.9} & \multicolumn{1}{c}{77.5} & \multicolumn{1}{c}{50.4} & \multicolumn{1}{c}{27.6} & \multicolumn{1}{c}{34.9} & \multicolumn{1}{c}{19.9} & \multicolumn{1}{c}{14.8} & \multicolumn{1}{c}{4.8}  \bigstrut[t]\\
     	\cline{3-12} & Market1501 &\multicolumn{1}{c}{14.3} & \multicolumn{1}{c}{14.6} & \multicolumn{1}{c}{73.9} & \multicolumn{1}{c}{71.2} & \multicolumn{1}{c}{92.6} & \multicolumn{1}{c}{81.8} & \multicolumn{1}{c}{48.9} & \multicolumn{1}{c}{29.5} & \multicolumn{1}{c}{19.1} & \multicolumn{1}{c}{6.5}  \bigstrut[t]\\
     	\cline{3-12} & Duke &\multicolumn{1}{c}{8.8} & \multicolumn{1}{c}{8.8} & \multicolumn{1}{c}{63.6} & \multicolumn{1}{c}{60.7} & \multicolumn{1}{c}{57.1} & \multicolumn{1}{c}{28.2} & \multicolumn{1}{c}{85.8} & \multicolumn{1}{c}{71.4} & \multicolumn{1}{c}{21.7} & \multicolumn{1}{c}{6.9}  \bigstrut[t]\\
     	\cline{3-12} & MSMT17 &\multicolumn{1}{c}{14.7} & \multicolumn{1}{c}{15.1} & \multicolumn{1}{c}{76.0} & \multicolumn{1}{c}{72.8} & \multicolumn{1}{c}{59.7} & \multicolumn{1}{c}{31.6} & \multicolumn{1}{c}{60.8} & \multicolumn{1}{c}{41.4} & \multicolumn{1}{c}{74.5} & \multicolumn{1}{c}{48.4}  \bigstrut[t]\\
     	\cline{3-12} & Joint Training &\multicolumn{1}{c}{66.8} & \multicolumn{1}{c}{64.5} & \multicolumn{1}{c}{89.6} & \multicolumn{1}{c}{88.3} & \multicolumn{1}{c}{93.1} & \multicolumn{1}{c}{81.8} & \multicolumn{1}{c}{84.5} & \multicolumn{1}{c}{72.7} & \multicolumn{1}{c}{74.6} & \multicolumn{1}{c}{49.2}  \bigstrut[t]\\    	
 
    	\hline
     	\multicolumn{1}{l}{\multirow{6}{*}{BoT-SNR*~\cite{SNR_cpvr2020}}} & CUHK03(L) &\multicolumn{1}{c}{71.9} & \multicolumn{1}{c}{70.6} & \multicolumn{1}{c}{77.7} & \multicolumn{1}{c}{74.5} & \multicolumn{1}{c}{65.4} & \multicolumn{1}{c}{38.9} & \multicolumn{1}{c}{41.2} & \multicolumn{1}{c}{25.6} & \multicolumn{1}{c}{22.1} & \multicolumn{1}{c}{7.5}  \bigstrut[t]\\
     	\cline{3-12} & CUHK-SYSU &\multicolumn{1}{c}{6.9} & \multicolumn{1}{c}{8.4} & \multicolumn{1}{c}{88.4} & \multicolumn{1}{c}{87.3} & \multicolumn{1}{c}{61.6} & \multicolumn{1}{c}{37.6} & \multicolumn{1}{c}{39.6} & \multicolumn{1}{c}{25.2} & \multicolumn{1}{c}{20.4} & \multicolumn{1}{c}{8.2}  \bigstrut[t]\\
     	\cline{3-12} & Market1501 &\multicolumn{1}{c}{17.6} & \multicolumn{1}{c}{17.5} & \multicolumn{1}{c}{79.7} & \multicolumn{1}{c}{77.2} & \multicolumn{1}{c}{95.0} & \multicolumn{1}{c}{86.4} & \multicolumn{1}{c}{49.7} & \multicolumn{1}{c}{32.4} & \multicolumn{1}{c}{21.1} & \multicolumn{1}{c}{7.6}  \bigstrut[t]\\
     	\cline{3-12} & Duke &\multicolumn{1}{c}{11.0} & \multicolumn{1}{c}{10.9} & \multicolumn{1}{c}{71.3} & \multicolumn{1}{c}{68.1} & \multicolumn{1}{c}{62.1} & \multicolumn{1}{c}{33.0} & \multicolumn{1}{c}{89.1} & \multicolumn{1}{c}{77.5} & \multicolumn{1}{c}{26.7} & \multicolumn{1}{c}{9.1}  \bigstrut[t]\\
     	\cline{3-12} & MSMT17 &\multicolumn{1}{c}{16.9} & \multicolumn{1}{c}{17.7} & \multicolumn{1}{c}{81.3} & \multicolumn{1}{c}{78.5} & \multicolumn{1}{c}{64.4} & \multicolumn{1}{c}{37.4} & \multicolumn{1}{c}{63.4} & \multicolumn{1}{c}{45.6} & \multicolumn{1}{c}{77.3} & \multicolumn{1}{c}{52.3}  \bigstrut[t]\\
     	\cline{3-12} & Joint Training &\multicolumn{1}{c}{66.6} & \multicolumn{1}{c}{65.8} & \multicolumn{1}{c}{90.7} & \multicolumn{1}{c}{89.5} & \multicolumn{1}{c}{94.2} & \multicolumn{1}{c}{84.9} & \multicolumn{1}{c}{87.5} & \multicolumn{1}{c}{76.4} & \multicolumn{1}{c}{76.2} & \multicolumn{1}{c}{51.3}  \bigstrut[t]\\    	
     	
    	\hline
     	\multicolumn{1}{l}{\multirow{6}{*}{BoT-DCSD (Ours)}} & CUHK03(L) &\multicolumn{1}{c}{66.8} & \multicolumn{1}{c}{66.5} & \multicolumn{1}{c}{75.5} & \multicolumn{1}{c}{72.3} & \multicolumn{1}{c}{49.1} & \multicolumn{1}{c}{25.4} & \multicolumn{1}{c}{24.2} & \multicolumn{1}{c}{13.3} & \multicolumn{1}{c}{10.3} & \multicolumn{1}{c}{3.3}  \bigstrut[t]\\
     	\cline{3-12} & CUHK-SYSU &\multicolumn{1}{c}{4.0} & \multicolumn{1}{c}{4.9} & \multicolumn{1}{c}{87.2} & \multicolumn{1}{c}{85.6} & \multicolumn{1}{c}{46.1} & \multicolumn{1}{c}{24.1} & \multicolumn{1}{c}{27.1} & \multicolumn{1}{c}{15.1} & \multicolumn{1}{c}{13.2} & \multicolumn{1}{c}{5.0}  \bigstrut[t]\\
     	\cline{3-12} & Market1501 &\multicolumn{1}{c}{7.6} & \multicolumn{1}{c}{8.0} & \multicolumn{1}{c}{76.7} & \multicolumn{1}{c}{87.8} & \multicolumn{1}{c}{95.3} & \multicolumn{1}{c}{87.8} & \multicolumn{1}{c}{32.0} & \multicolumn{1}{c}{18.9} & \multicolumn{1}{c}{10.2} & \multicolumn{1}{c}{3.4}  \bigstrut[t]\\
     	\cline{3-12} & Duke &\multicolumn{1}{c}{5.4} & \multicolumn{1}{c}{5.6} & \multicolumn{1}{c}{66.9} & \multicolumn{1}{c}{63.5} & \multicolumn{1}{c}{48.6} & \multicolumn{1}{c}{22.3} & \multicolumn{1}{c}{88.6} & \multicolumn{1}{c}{79.0} & \multicolumn{1}{c}{14.1} & \multicolumn{1}{c}{4.5}  \bigstrut[t]\\
     	\cline{3-12} & MSMT17 &\multicolumn{1}{c}{12.5} & \multicolumn{1}{c}{12.3} & \multicolumn{1}{c}{78.3} & \multicolumn{1}{c}{75.0} & \multicolumn{1}{c}{52.4} & \multicolumn{1}{c}{27.1} & \multicolumn{1}{c}{49.9} & \multicolumn{1}{c}{33.1} & \multicolumn{1}{c}{79.8} & \multicolumn{1}{c}{59.5}  \bigstrut[t]\\
     	\cline{3-12} & Joint Training &\multicolumn{1}{c}{\textbf{79.4}} & \multicolumn{1}{c}{\textbf{78.0}} & \multicolumn{1}{c}{\textbf{92.4}} & \multicolumn{1}{c}{\textbf{91.3}} & \multicolumn{1}{c}{\textbf{95.9}} & \multicolumn{1}{c}{\textbf{89.5}} & \multicolumn{1}{c}{\textbf{89.6}} & \multicolumn{1}{c}{\textbf{80.6}} & \multicolumn{1}{c}{\textbf{80.3}} & \multicolumn{1}{c}{\textbf{61.5}}  \bigstrut[t]\\    	 
    	
    	\hline \hline
    \end{tabular}
    
\end{table*}

\begin{figure*}[t]
\begin{center}
\resizebox{1.0\textwidth}{!}{
\includegraphics[width=\linewidth]{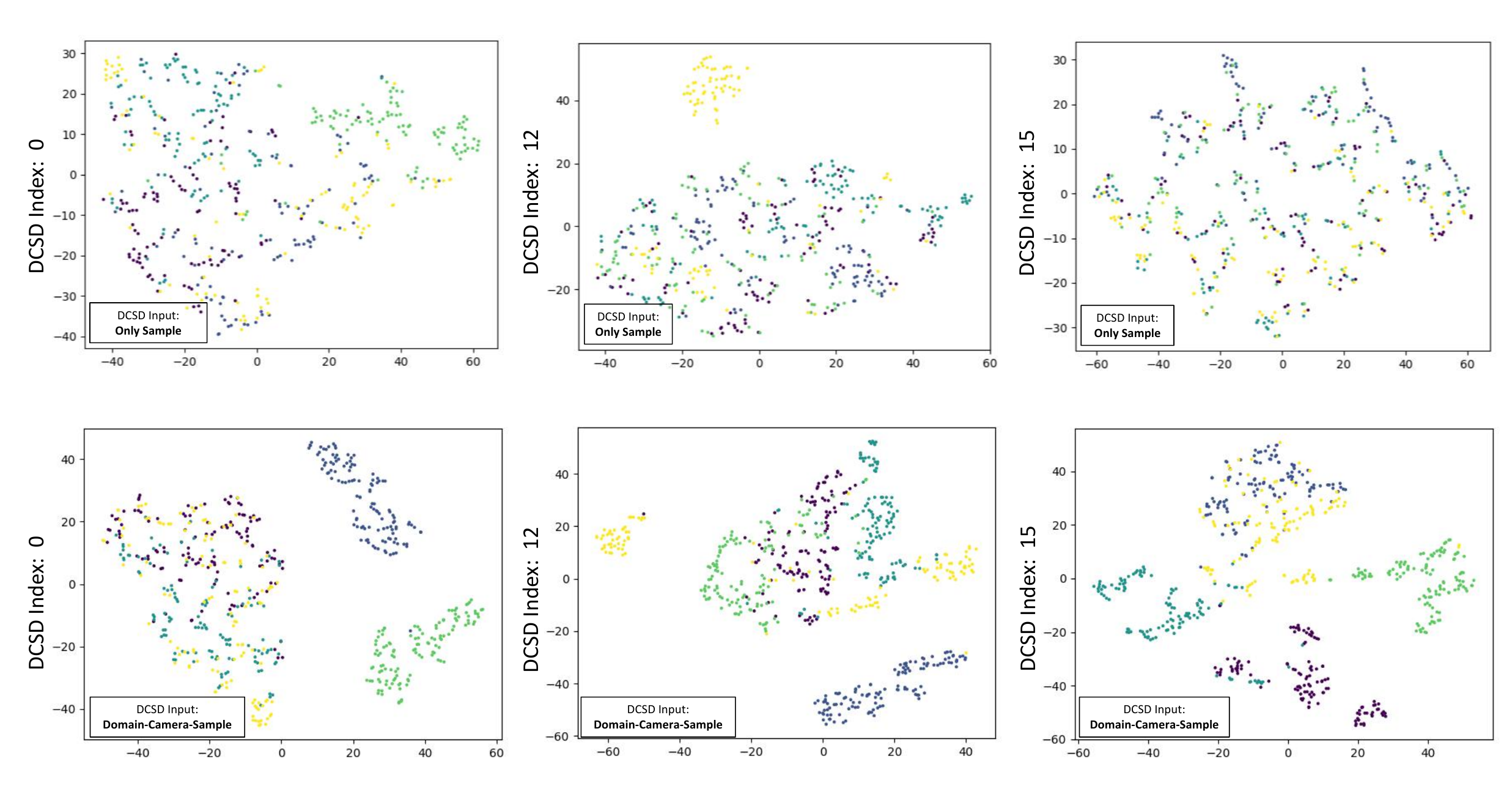}
}
\caption{
Visualization of the generated parameter in DCSD. There are $16$ DCSD modules in our ResNet50-DCSD, and we visualize the generated parameters of $0th$, $12th$, and $15th$ DCSD. The first column shows the visualization of convolution parameters generated only by sample features, and the second column shows the visualization of convolution parameters generated by comprehensively considering domain, camera, and sample information. The same color represents the same domain, best viewed in color.
}
\label{fig:TSNE}
\end{center}
\end{figure*}

\section{Experiments}

\begin{table}[]
\caption{The Params and FLOPs comparison with ResNet50.}
\label{tab:Model}
\centering
\begin{tabular}{ccc}
\hline \hline
Model         & Params & FLOPs \\
\hline
ResNet50      & 25.6M  & 4.1B  \\
ResNet50-DCSD & 18.2M  & 2.3B \\   
\hline \hline
\end{tabular}
\end{table}

\subsection{Datasets}
We conduct extensive experiments on five public person ReID datasets, $i.e.$, Market1501, CUHK-SYSU, DukeMTMC, CUHK03 and MSMT17. The detailed statistics of person Re-Identification datasets are shown in Table~\ref{tab:datasets}. For the joint training of multiple datasets, we mix the five datasets together and take random samples for training. The Cumulative Match Curve (CMC) at Rank-1 and the mean Average Precision (mAP) are used as the evaluation criteria.

\noindent{\bf{Market1501}}~\cite{iccv15zheng} contains $32,668$ images of $1,501$ labeled persons of six camera views. It is collected from Tsinghua University and the pictures are from the outdoors. There are $751$ identities in the training set and $750$ identities in the testing set. View overlapping exists among different cameras, including $5$ high-resolution cameras, and a low-resolution camera.

\noindent{\bf{CUHK-SYSU}}~\cite{2017Joint} were originally used for person search which is collected from street snap and movies. And we modified the original dataset by using the ground-truth person bounding box annotation. For testing on this dataset, we fixed both query and gallery sets instead of using variable gallery sets. We used 2,900 query persons, with each query containing at least one image in the gallery. Note that CUHK-SYSU lacks camera ID information, so the cross-camera setting is ignored during the evaluation phase.

\noindent{\bf{Duke}}~\cite{iccv17duke} is a subset of the DukeMTMC used for person re-identification by images.
It is collected from Duke University and the pictures are from the outdoors.
It consists of $36,411$ images of $1,812$ persons from 8 high-resolution cameras. $16,522$
images of $702$ persons are randomly selected from the dataset as the training set, and the remaining $702$ persons are divided into the testing set where contains $2,228$ query images and $17,661$ gallery
images.

\noindent{\bf{CUHK03}}~\cite{cvpr14cuhk} is collected from The Chinese University of Hong Kong and the pictures are from the indoors and outdoors.
It has two ways of annotating bounding box including labelled by humans or detected from deformable part models (DPMs). The labeled dataset includes $7,368$ training, $1,400$ query and $5,328$ gallery images while detected dataset consists of $7,365$ training, $1,400$ query and $5,332$ gallery images. We use the labeled dataset as default.

\noindent{\bf{MSMT17}}~\cite{cvpr18msmt} is the current largest publicly available person Re-ID dataset. It has $126,441$ images of $4,101$ identities from indoors and outdoors. The video is collected with different weather conditions at three-time slots (morning, noon, afternoon). All annotations, including camera IDs, weathers, and time slots, are available.

\subsection{Implementation Details}
The proposed model is implemented using PyTorch in Ubuntu16.04. All experiments are conducted on an NVIDIA GTX 1080Ti GPU with 11GB memory. The images were resized to $256 \times 128$ and the batch size is $64$ ($4$ images/ID and $16$ IDs).
We didn't utilize the test with flip and re-ranking.
Unless specified otherwise, we use BoT~\cite{luo2019bag} as the baseline by default. The backbone is pre-trained on ImageNet.
The margin in the triplet loss is $0.3$ in all our experiments.
The model has trained $120$ epochs and the learning rate is initialized to $3.5\times 10^{-4}$ and divided by $10$ at the $40$th epoch and $90$th epoch.
The detailed implementation of other methods follows the settings in their respective papers.

\subsection{Experimental Results}
In this section, we evaluate our proposed DCSD in comparison with state-of-the-art approaches on several benchmarks.
For a fair comparison, we only demonstrate the performance of the TransReID~\cite{he2021transreid} model trained with DeiT-S/16 which is comparable with ResNet50.

It can be seen from Table~\ref{tab:SOTA_person} that after the joint training of multiple datasets (domains), 
the ReID performance of BoT and AGW decreased significantly on most of the datasets. Although the data involved in training increased, the performance decreased due to the existence of multi-domain conflicts. For example, the Rank-1 of AGW on CUHK03 drop from $75.1\%$ to $70.3\%$. 

When we replace the ResNet50 in BoT and AGW with its corresponding DCSD model, the performance has increased on some datasets, such as MSMT17 and Duke. And the performance dropped on CUHK-SYSU. The model size and FLOPs are listed in Table~\ref{tab:Model}.
The reason for this phenomenon may be that MSMT17 and Duke have many cameras ($15$ and $8$), so there may be some sub-domain conflict inside MSMT17 and Duke, so ``BoT-DCSD'' and ``AGW-DCSD'' can get good performance on MSMT17 and Duke.
However, CUHK-SYSU only has one camera, Therefore, the performance with DCSD on CUHK-SYSU is not so well.
For DCSD joint training, consistent performance improvements were achieved for all datasets. 
Note that the mAP of ``BoT-DCSD'' on CUHK03 and CUHK-SYSU are lower than that of ``BoT'' ($-0.8\% and -0.4\%$), but the mAP of ``BoT-DCSD (joint training)'' are higher than that of ``BoT (joint training)'' ($+12.3\% and +2.0\%$).
Compared with the original method of joint training, ``Joint Training Gain'' can get up to $12.3\%$ performance (mAP) improvement. Compared with other approaches, the 
performance of our method achieves state-of-the-art performance on all of the datasets.

 \subsection{Comparison with Domain Generalization Methods}
Domain generalization~\cite{dual_norm_bmvc2019,SNR_cpvr2020,MetaBIN_CVPR_2021} has attracted much attention in person ReID recently. It aims to make a model trained on multiple source domains generalize to an unseen target domain which can alleviate the domain discrepancy between datasets. In this subsection, we experiment whether the traditional domain generalization methods can solve the problem of domain conflict in multi-domain joint training for person ReID.
We use BoT~\cite{luo2019bag} as the baseline, and then add DN~\cite{dual_norm_bmvc2019}, SNR~\cite{SNR_cpvr2020} or DCSD (ours) to the BoT. 
Table~\ref{tab:DG} shows that all three methods show better domain generalization ability than the baseline.
However, when multiple domains are jointly trained, the performance of BoT-DN and BoT-SNR are even worse than that of each domain independently trained. Only BoT-DCSD (ours) can achieve consistent performance improvement on all of the benchmarks.
One possible reason is that learning a domain agnostic model is difficult, since different domains can give rise to very different image distributions~\cite{li2021dynamic}. When forcing a model to be domain agnostic, it essentially averages the domain conflict rather than avoids it.
Therefore, the domain generalization methods may not work well in multi-domain joint training.

\section{Ablation Study}
In this section, we conduct investigations on how several key factors affect the model’s overall performance. 

\subsection{Domain-Related Factors}
We divide the domain-related factors into internal domain-related factor and external domain-related factors.
Specifically, we regard the domain information and camera information as the external domain-related factors while sample information as the internal domain-related factor. 
And in this sub-section, we study the impact of these domain-related factors on performance in multi-domain joint training.
In Table~\ref{tab:factors}, we can see that the performance of multi-domain joint training is low when DCSD is not used. With the gradual increase of using sample information, domain information and camera information to dynamically generate convolution parameters, its performance began to improve gradually. Finally, the mAP improved by $2.0\%$ to $12.3\%$ on different datasets.
The experimental results show that both internal and external domain-related factors are helpful to generate dynamic convolution parameters, so as to reduce domain conflict.

%

\begin{table}[th]
\caption{The performance (mAP) with the domain-related factors to  generate the sample-specific network in DCSD.  ``D'', ``C'', ``S'' and ``JT'' are short for ``Domain Information'', ``Camera Information'', ``Sample Information'' and ``Joint Training'' respectively. .
}
\label{tab:factors}
\centering
\scalebox{0.73}{
\begin{tabular}{lccccc}
\hline \hline
Method & CUHK03(L) & CUHK-SYSU & Market1501 & Duke & MSMT17 \\ \hline
BoT (JT)                 &  65.7  &  89.3  &  84.6  &  75.9  &  49.5  \\
BoT-DCSD (JT), S         &  74.6  &  90.9  &  88.2  &  80.2  &  58.8  \\
BoT-DCSD (JT), S + D     &  75.2  &  91.0  &  88.5  &  80.4  &  59.6  \\
BoT-DCSD (JT), S + C     &  77.9  &  91.1  &  89.1  &  80.4  &  61.2  \\
BoT-DCSD (JT), S + D + C &\textbf{78.0}&\textbf{91.3}&\textbf{89.5}&\textbf{80.6} &\textbf{61.5}\\ \hline \hline
\end{tabular} 
}
\end{table}

\subsection{Stop Gradient}
It can be seen from Figure~\ref{fig:architecture} that the camera information and domain information predicted by MLPs will be used in two branches: the first branch is to predict the corresponding IDs and the second branch is to generate convolution parameters. 
In order to make the camera information and domain information more pure, we cut off the gradient propagation on the second branch that the camera information and domain information are only supervised by the corresponding IDs. We named it Stop Gradient (SG).
In Table~\ref{tab:StopGradient} we can see that compared with ``DCSD without GS'', ``DCSD with GS'' achieve consistent performance improvement on all datasets.

\begin{table}[th]
\caption{DCSD with or without the Stop Gradient (SG). ``JT'' is short for ``Joint Training''.}
\label{tab:StopGradient}
\centering
\scalebox{0.73}{
\begin{tabular}{lccccc}
\hline \hline
Method & CUHK03(L) & CUHK-SYSU & Market1501 & Duke & MSMT17 \\ \hline
BoT-DCSD (JT) w/o GS &  76.0  &  90.5  &  88.9 &  80.4  &  61.0  \\
BoT-DCSD (JT) with GS &  78.0  &  91.3  &  89.5  &  80.6 &  61.5 \\ \hline
AGW-DCSD (JT) w/o GS &  76.3  &  90.9  &  88.9 &  80.9  &  61.4  \\
AGW-DCSD (JT) with GS &  78.8  &  91.4  &  89.3  &  81.2  &  62.9\\ 
\hline \hline
\end{tabular}
}
\end{table}

\subsection{Visualization of Generated Parameter}
We visualize the parameters generated by samples from different domains by t-SNE~\cite{maaten2008visualizing}.
T-SNE visualizes high-dimensional data by giving each datapoint a location in a two or three-dimensional map. It is a variation of stochastic neighbor embedding that is much easier to optimize, and produces significantly better visualizations by reducing the tendency to crowd points together in the center of the map.
In Figure~\ref{fig:TSNE}, we can see that if only sample information is used to generate convolution parameters, the parameters generated by samples from different domains are not very different. But the parameters generated by the combination of domain, camera and sample information have good domain discrimination. It means that the network can dynamically generate a sample specific network which is suitable for the current domain.

\IEEEpeerreviewmaketitle
 
\section{Conclusion}
In this work, we deliver some surprising findings that the traditional static network is tough to deal with the conflict between multiple domains for person re-identification, which may lead to arousing the performance degradation when multi-domain joint training.
And then we propose a Domain-Camera-Sample Dynamic network (DCSD) whose parameters can be adaptive to various factors.
Without the need to dealing with the complex relationship between different domains,
DCSD uses internal domain-related factor, such as features from the sample information, and external domain-related factors, such as domain information and camera information, to dynamically generate the sample-specific network.
Experimental results show that DCSD can boost the performance (up to $12.3\%$) while joint training in multiple domains.

\ifCLASSOPTIONcaptionsoff
  \newpage
\fi

\bibliographystyle{IEEEtran}
\bibliography{IEEEabrv,sample-base}

\begin{thebibliography}{10}
\providecommand{\url}[1]{#1}
\csname url@samestyle\endcsname
\providecommand{\newblock}{\relax}
\providecommand{\bibinfo}[2]{#2}
\providecommand{\BIBentrySTDinterwordspacing}{\spaceskip=0pt\relax}
\providecommand{\BIBentryALTinterwordstretchfactor}{4}
\providecommand{\BIBentryALTinterwordspacing}{\spaceskip=\fontdimen2\font plus
\BIBentryALTinterwordstretchfactor\fontdimen3\font minus
  \fontdimen4\font\relax}
\providecommand{\BIBforeignlanguage}[2]{{%
\expandafter\ifx\csname l@#1\endcsname\relax
\typeout{** WARNING: IEEEtran.bst: No hyphenation pattern has been}%
\typeout{** loaded for the language `#1'. Using the pattern for}%
\typeout{** the default language instead.}%
\else
\language=\csname l@#1\endcsname
\fi
#2}}
\providecommand{\BIBdecl}{\relax}
\BIBdecl

\bibitem{krizhevsky2012imagenet}
A.~Krizhevsky, I.~Sutskever, and G.~E. Hinton, ``Imagenet classification with
  deep convolutional neural networks,'' \emph{Advances in neural information
  processing systems}, vol.~25, pp. 1097--1105, 2012.

\bibitem{he2016deep}
K.~He, X.~Zhang, S.~Ren, and J.~Sun, ``Deep residual learning for image
  recognition,'' in \emph{Proceedings of the IEEE conference on computer vision
  and pattern recognition}, 2016, pp. 770--778.

\bibitem{sun2017revisiting}
C.~Sun, A.~Shrivastava, S.~Singh, and A.~Gupta, ``Revisiting unreasonable
  effectiveness of data in deep learning era,'' in \emph{Proceedings of the
  IEEE international conference on computer vision}, 2017, pp. 843--852.

\bibitem{Qi_2019_ICCV}
L.~Qi, L.~Wang, J.~Huo, L.~Zhou, Y.~Shi, and Y.~Gao, ``A novel unsupervised
  camera-aware domain adaptation framework for person re-identification,'' in
  \emph{Proceedings of the IEEE/CVF International Conference on Computer Vision
  (ICCV)}, October 2019.

\bibitem{li2021dynamic}
Y.~Li, L.~Yuan, Y.~Chen, P.~Wang, and N.~Vasconcelos, ``Dynamic transfer for
  multi-source domain adaptation,'' in \emph{Proceedings of the IEEE/CVF
  Conference on Computer Vision and Pattern Recognition}, 2021, pp.
  10\,998--11\,007.

\bibitem{ganin2015unsupervised}
Y.~Ganin and V.~Lempitsky, ``Unsupervised domain adaptation by
  backpropagation,'' in \emph{International conference on machine
  learning}.\hskip 1em plus 0.5em minus 0.4em\relax PMLR, 2015, pp. 1180--1189.

\bibitem{2016Return}
B.~Sun, J.~Feng, and K.~Saenko, ``Return of frustratingly easy domain
  adaptation,'' 2016.

\bibitem{sun2016deep}
B.~Sun and K.~Saenko, ``Deep coral: Correlation alignment for deep domain
  adaptation,'' in \emph{European conference on computer vision}.\hskip 1em
  plus 0.5em minus 0.4em\relax Springer, 2016, pp. 443--450.

\bibitem{bousmalis2017unsupervised}
K.~Bousmalis, N.~Silberman, D.~Dohan, D.~Erhan, and D.~Krishnan, ``Unsupervised
  pixel-level domain adaptation with generative adversarial networks,'' in
  \emph{Proceedings of the IEEE conference on computer vision and pattern
  recognition}, 2017, pp. 3722--3731.

\bibitem{liu2016coupled}
M.-Y. Liu and O.~Tuzel, ``Coupled generative adversarial networks,''
  \emph{Advances in neural information processing systems}, vol.~29, pp.
  469--477, 2016.

\bibitem{hoffman2018cycada}
J.~Hoffman, E.~Tzeng, T.~Park, J.-Y. Zhu, P.~Isola, K.~Saenko, A.~Efros, and
  T.~Darrell, ``Cycada: Cycle-consistent adversarial domain adaptation,'' in
  \emph{International conference on machine learning}.\hskip 1em plus 0.5em
  minus 0.4em\relax PMLR, 2018, pp. 1989--1998.

\bibitem{Bai_2021_CVPR}
Z.~Bai, Z.~Wang, J.~Wang, D.~Hu, and E.~Ding, ``Unsupervised multi-source
  domain adaptation for person re-identification,'' in \emph{Proceedings of the
  IEEE/CVF Conference on Computer Vision and Pattern Recognition (CVPR)}, June
  2021, pp. 12\,914--12\,923.

\bibitem{2020Dynamic}
Y.~Chen, X.~Dai, M.~Liu, D.~Chen, and Z.~Liu, ``Dynamic convolution: Attention
  over convolution kernels,'' in \emph{2020 IEEE/CVF Conference on Computer
  Vision and Pattern Recognition (CVPR)}, 2020.

\bibitem{zhang2020dynet}
Y.~Zhang, J.~Zhang, Q.~Wang, and Z.~Zhong, ``Dynet: Dynamic convolution for
  accelerating convolutional neural networks,'' 2020.

\bibitem{Li_2021_CVPR}
D.~Li, J.~Hu, C.~Wang, X.~Li, Q.~She, L.~Zhu, T.~Zhang, and Q.~Chen,
  ``Involution: Inverting the inherence of convolution for visual
  recognition,'' in \emph{Proceedings of the IEEE/CVF Conference on Computer
  Vision and Pattern Recognition (CVPR)}, June 2021, pp. 12\,321--12\,330.

\bibitem{Zhou_2021_CVPR}
J.~Zhou, V.~Jampani, Z.~Pi, Q.~Liu, and M.-H. Yang, ``Decoupled dynamic filter
  networks,'' in \emph{Proceedings of the IEEE/CVF Conference on Computer
  Vision and Pattern Recognition (CVPR)}, June 2021, pp. 6647--6656.

\bibitem{sun2017beyond}
Y.~Sun, L.~Zheng, Y.~Yang, Q.~Tian, and S.~Wang, ``Beyond part models: Person
  retrieval with refined part pooling (and a strong convolutional baseline),''
  in \emph{ECCV}, 2018, pp. 480--496.

\bibitem{almazan2018re}
J.~Almazan, B.~Gajic, N.~Murray, and D.~Larlus, ``Re-id done right: towards
  good practices for person re-identification,'' \emph{arXiv:1801.05339}, 2018.

\bibitem{song2018mask}
C.~Song, Y.~Huang, W.~Ouyang, and L.~Wang, ``Mask-guided contrastive attention
  model for person re-identification,'' in \emph{CVPR}, 2018, pp. 1179--1188.

\bibitem{qi2018maskreid}
L.~Qi, J.~Huo, L.~Wang, Y.~Shi, and Y.~Gao, ``Maskreid: A mask based deep
  ranking neural network for person re-identification,''
  \emph{arXiv:1804.03864}, 2018.

\bibitem{xu2018attention}
J.~Xu, R.~Zhao, F.~Zhu, H.~Wang, and W.~Ouyang, ``Attention-aware compositional
  network for person re-identification,'' in \emph{CVPR}, 2018, pp. 2119--2128.

\bibitem{kalayeh2018human}
M.~M. Kalayeh, E.~Basaran, M.~G{\"o}kmen, M.~E. Kamasak, and M.~Shah, ``Human
  semantic parsing for person re-identification,'' in \emph{CVPR}, 2018, pp.
  1062--1071.

\bibitem{li2018harmonious}
W.~Li, X.~Zhu, and S.~Gong, ``Harmonious attention network for person
  re-identification,'' in \emph{CVPR}, 2018, pp. 2285--2294.

\bibitem{si2018dual}
J.~Si, H.~Zhang, C.-G. Li, J.~Kuen, X.~Kong, A.~C. Kot, and G.~Wang, ``Dual
  attention matching network for context-aware feature sequence based person
  re-identification,'' in \emph{CVPR}, 2018, pp. 5363--5372.

\bibitem{wang2018mancs}
C.~Wang, Q.~Zhang, C.~Huang, W.~Liu, and X.~Wang, ``Mancs: A multi-task
  attentional network with curriculum sampling for person re-identification,''
  in \emph{ECCV}, 2018, pp. 365--381.

\bibitem{wang2018learning}
G.~Wang, Y.~Yuan, X.~Chen, J.~Li, and X.~Zhou, ``Learning discriminative
  features with multiple granularities for person re-identification,'' in
  \emph{ACM Multimedia}, 2018, pp. 274--282.

\bibitem{fu2019horizontal}
Y.~Fu, Y.~Wei, Y.~Zhou, H.~Shi, G.~Huang, X.~Wang, Z.~Yao, and T.~Huang,
  ``Horizontal pyramid matching for person re-identification,'' in \emph{AAAI},
  vol.~33, 2019, pp. 8295--8302.

\bibitem{zhang2019densely}
Z.~Zhang, C.~Lan, W.~Zeng, and Z.~Chen, ``Densely semantically aligned person
  re-identification,'' in \emph{CVPR}, 2019, pp. 667--676.

\bibitem{2019Omni}
K.~Zhou, Y.~Yang, A.~Cavallaro, and T.~Xiang, ``Omni-scale feature learning for
  person re-identification,'' 2019.

\bibitem{2020IANet}
R.~Hou, B.~Ma, H.~Chang, X.~Gu, S.~Shan, and X.~Chen,
  ``Interaction-and-aggregation network for person re-identification,'' in
  \emph{CVPR}, 2020.

\bibitem{wang2020high}
G.~Wang, S.~Yang, H.~Liu, Z.~Wang, Y.~Yang, S.~Wang, G.~Yu, E.~Zhou, and
  J.~Sun, ``High-order information matters: Learning relation and topology for
  occluded person re-identification,'' in \emph{CVPR}, 2020, pp. 6449--6458.

\bibitem{zhu2020identity}
K.~Zhu, H.~Guo, Z.~Liu, M.~Tang, and J.~Wang, ``Identity-guided human semantic
  parsing for person re-identification,'' \emph{arXiv:2007.13467}, 2020.

\bibitem{he2021transreid}
S.~He, H.~Luo, P.~Wang, F.~Wang, H.~Li, and W.~Jiang, ``Transreid:
  Transformer-based object re-identification,'' 2021.

\bibitem{luo2019bag}
H.~Luo, W.~Jiang, Y.~Gu, F.~Liu, X.~Liao, S.~Lai, and J.~Gu, ``A strong
  baseline and batch normalization neck for deep person re-identification,''
  \emph{TMM}, 2019.

\bibitem{2021AGW}
M.~Ye, J.~Shen, G.~Lin, T.~Xiang, and S.~Hoi, ``Deep learning for person
  re-identification: A survey and outlook,'' \emph{IEEE Transactions on Pattern
  Analysis and Machine Intelligence}, vol.~PP, no.~99, pp. 1--1, 2021.

\bibitem{dual_norm_bmvc2019}
\BIBentryALTinterwordspacing
J.~Jia, Q.~Ruan, and T.~Hospedales, ``Frustratingly easy person
  re-identification: Generalizing person re-id in practice,'' in
  \emph{Proceedings of the British Machine Vision Conference (BMVC)},
  K.~Sidorov and Y.~Hicks, Eds.\hskip 1em plus 0.5em minus 0.4em\relax BMVA
  Press, September 2019, pp. 141.1--141.14. [Online]. Available:
  \url{https://dx.doi.org/10.5244/C.33.141}
\BIBentrySTDinterwordspacing

\bibitem{SNR_cpvr2020}
X.~Jin, C.~Lan, W.~Zeng, Z.~Chen, and L.~Zhang, ``Style normalization and
  restitution for generalizable person re-identification,'' in \emph{2020
  IEEE/CVF Conference on Computer Vision and Pattern Recognition (CVPR)}, 2020.

\bibitem{iccv15zheng}
L.~Zheng, L.~Shen, L.~Tian, S.~Wang, J.~Wang, and Q.~Tian, ``Scalable person
  re-identification: A benchmark,'' in \emph{ICCV}, 2015, pp. 1116--1124.

\bibitem{2017Joint}
X.~Tong, L.~Shuang, B.~Wang, L.~Liang, and X.~Wang, ``Joint detection and
  identification feature learning for person search,'' in \emph{2017 IEEE
  Conference on Computer Vision and Pattern Recognition (CVPR)}, 2017.

\bibitem{iccv17duke}
Z.~Zheng, L.~Zheng, and Y.~Yang, ``Unlabeled samples generated by gan improve
  the person re-identification baseline in vitro,'' in \emph{ICCV}, 2017, pp.
  3754--3762.

\bibitem{cvpr14cuhk}
W.~Li, R.~Zhao, T.~Xiao, and X.~Wang, ``Deepreid: Deep filter pairing neural
  network for person re-identification,'' in \emph{CVPR}, 2014, pp. 152--159.

\bibitem{cvpr18msmt}
L.~Wei, S.~Zhang, W.~Gao, and Q.~Tian, ``Person transfer gan to bridge domain
  gap for person re-identification,'' in \emph{CVPR}, 2018, pp. 79--88.

\bibitem{MetaBIN_CVPR_2021}
S.~Choi, T.~Kim, M.~Jeong, H.~Park, and C.~Kim, ``Meta batch-instance
  normalization for generalizable person re-identification,'' in
  \emph{Proceedings of the IEEE/CVF Conference on Computer Vision and Pattern
  Recognition (CVPR)}, June 2021, pp. 3425--3435.

\bibitem{maaten2008visualizing}
L.~v.~d. Maaten and G.~Hinton, ``Visualizing data using t-sne,'' \emph{Journal
  of machine learning research}, vol.~9, no. Nov, pp. 2579--2605, 2008.

\end{thebibliography}

\end{document}